\documentclass[11pt,a4paper]{article}

\usepackage[utf8]{inputenc}
\usepackage[T1]{fontenc}
\usepackage{times}
\usepackage{graphicx}
\usepackage{booktabs}
\usepackage{multirow}
\usepackage{colortbl}
\usepackage{amsmath}
\usepackage{hyperref}
\usepackage{xcolor}
\usepackage{tikz}
\usepackage{pgfplots}
\usepackage[margin=2.5cm]{geometry}
\usepackage{float}
\usepackage{subcaption}
\usepackage{natbib}
\usepackage{authblk}

\pgfplotsset{compat=1.18}
\usetikzlibrary{shapes, arrows.meta, positioning, calc, patterns}

\title{\textbf{IndoBERT-Sentiment: Context-Conditioned Sentiment\\Classification for Indonesian Text}}

\author[1]{Muhammad Apriandito Arya Saputra}
\author[2]{Andry Alamsyah}
\author[2]{Dian Puteri Ramadhani}
\author[3]{Thomhert Suprapto Siadari}
\author[4]{Hanif Fakhrurroja}

\affil[1]{SocialX \\ \texttt{apriandito@socialx.id}}
\affil[2]{Center of Excellence SAKTI, Research Institute Intelligent Business \& Sustainable Economy, Telkom University \\ \texttt{\{andry, dianpramadhani\}@telkomuniversity.ac.id}}
\affil[3]{Biomedical Engineering Study Program, School of Electrical Engineering, Telkom University \\ \texttt{thomhert@telkomuniversity.ac.id}}
\affil[4]{Research Center for Smart Mechatronics, National Research and Innovation Agency (BRIN) \\ \texttt{hani010@brin.go.id}}

\date{}

\begin{document}

\maketitle

\begin{abstract}
Existing Indonesian sentiment analysis models classify text in isolation, ignoring the topical context that often determines whether a statement is positive, negative, or neutral. We introduce IndoBERT-Sentiment, a context-conditioned sentiment classifier that takes both a topical context and a text as input, producing sentiment predictions grounded in the topic being discussed. Built on IndoBERT Large (335M parameters) and trained on 31,360 context--text pairs labeled across 188 topics, the model achieves an F1 macro of 0.856 and accuracy of 88.1\%. In a head-to-head evaluation against three widely used general-purpose Indonesian sentiment models on the same test set, IndoBERT-Sentiment outperforms the best baseline by 35.6 F1 points. We show that context-conditioning, previously demonstrated for relevancy classification, transfers effectively to sentiment analysis and enables the model to correctly classify texts that are systematically misclassified by context-free approaches.
\end{abstract}

\section{Introduction}

Sentiment analysis is one of the most widely deployed NLP tasks, and for Bahasa Indonesia it is no exception. Several pre-trained sentiment classifiers are publicly available, collectively downloaded hundreds of thousands of times from model repositories. These models take a single text as input and produce a sentiment label: positive, negative, or neutral.

Yet this formulation has a fundamental limitation. Sentiment is not an intrinsic property of text; it is a relationship between text and a topic. The statement ``angkanya terus naik setiap bulan'' (``the numbers keep rising every month'') is negative when discussing inflation but positive when discussing economic growth. A statement like ``KPK tangkap bupati yang korupsi dana bansos'' (``KPK arrests regent who embezzled social aid funds'') is positive from the perspective of anti-corruption efforts, yet a context-free classifier sees only the words ``korupsi'' and ``tangkap'' and defaults to neutral.

This paper asks a simple question: \emph{does adding topical context to sentiment analysis improve performance?}

In prior work \citep{saputra2026indobert}, we introduced context-conditioned classification for relevancy, showing that a model taking \texttt{[CLS] context [SEP] text [SEP]} as input could determine whether a text is relevant to a given topic with F1 of 0.948. Here, we apply the same architecture and methodology to a different task: instead of predicting relevancy, we predict sentiment. The architecture is identical. The training data is the same set of context--text pairs, re-labeled for sentiment. The question is whether the approach transfers.

It does. Our model, IndoBERT-Sentiment, achieves an F1 macro of 0.856, compared to 0.487--0.501 for the three most widely used general-purpose Indonesian sentiment classifiers evaluated on the same test set. The improvement is particularly dramatic for positive sentiment, where baseline models achieve F1 scores of 0.135--0.211 while our model achieves 0.791.

Our contributions are:
\begin{enumerate}
    \item A demonstration that \textbf{context-conditioned classification transfers} from relevancy to sentiment, using the same architecture and the same dataset re-labeled for a new task.
    \item A \textbf{head-to-head benchmark} of four models on the same test set, showing that context-conditioning provides substantial improvements over context-free sentiment analysis.
    \item \textbf{Publicly available models} for context-conditioned Indonesian sentiment classification, in both three-class (Negatif/Netral/Positif) and binary (Negatif/Positif) variants.
\end{enumerate}

\section{Related Work}

\subsection{Indonesian Sentiment Analysis}

Indonesian sentiment analysis has benefited from the IndoNLU benchmark \citep{wilie2020indonlu}, which established standard datasets and baselines for several NLP tasks including sentiment classification. The SmSA (Sentiment Analysis) task within IndoNLU uses document-level product reviews with three labels (positive, negative, neutral), and most publicly available Indonesian sentiment models are fine-tuned on this dataset.

The three most downloaded Indonesian sentiment classifiers on HuggingFace are all fine-tuned on SmSA or similar review datasets: a BERT Base Indonesian model with approximately 108,000 downloads, a RoBERTa Base Indonesian model with approximately 80,000 downloads, and an IndoBERT Base model with approximately 15,000 downloads. All three take a single text as input and produce a sentiment label without considering any topical context.

\subsection{Context-Conditioned Classification}

The idea of conditioning text classification on an external context has been explored in several forms. Aspect-based sentiment analysis \citep{pontiki2014semeval} conditions sentiment on a specific aspect mentioned within the text. Targeted sentiment analysis \citep{mitchell2013open} predicts sentiment toward a named entity. Natural language inference \citep{bowman2015snli} determines the relationship between a premise and hypothesis.

Our approach is closest to targeted sentiment analysis but operates at the topic level rather than the entity level. The context is a short topic description (e.g., ``Pertumbuhan ekonomi Indonesia''), and the model predicts the sentiment of a given text \emph{with respect to that topic}. This formulation is particularly relevant for social media monitoring, where the same text may have different sentiment implications depending on which topic is being tracked.

\subsection{From Relevancy to Sentiment}

In \citet{saputra2026indobert}, we introduced IndoBERT-Relevancy, a context-conditioned classifier that determines whether a text is relevant to a given topic. That model was trained on 31,360 context--text pairs across 188 topics, using an iterative data construction process that progressively improved handling of formal text, informal text, and implicit references. The present work reuses that exact dataset and architecture, asking whether the same approach that proved effective for relevancy can be transferred to sentiment classification.

\section{Method}

\subsection{Architecture}

We use the same architecture as IndoBERT-Relevancy: IndoBERT Large P2 \citep{wilie2020indonlu} (335 million parameters) with a classification head. The input is formatted as:

\begin{center}
\texttt{[CLS] context [SEP] text [SEP]}
\end{center}

The \texttt{[CLS]} representation is mapped to three output classes: Negatif (0), Netral (1), and Positif (2).

\subsection{Dataset}

We reuse the 31,360 context--text pairs from IndoBERT-Relevancy. These pairs span 188 topical contexts across 12 thematic domains and include three types of text: formal news headlines (18,798 pairs), informal social media posts (7,884 pairs), and LLM-generated implicit references (4,678 pairs).

The original dataset was labeled for binary relevancy. We re-labeled all 31,360 pairs for sentiment using GPT-4o-mini \citep{openai2024gpt4o} with the following instruction: given a context (topic description) and a text, determine whether the text expresses positive, neutral, or negative sentiment \emph{toward the topic described by the context}. The labeling was performed with temperature 0.0 and structured JSON output, achieving 72.6\% high confidence, 26.8\% medium confidence, and 0.6\% low confidence.

The resulting sentiment distribution is shown in Table~\ref{tab:data}.

\begin{table}[H]
\centering
\caption{Dataset statistics after re-labeling for sentiment.}
\label{tab:data}
\begin{tabular}{lrr}
\toprule
\textbf{Sentiment} & \textbf{Count} & \textbf{Percentage} \\
\midrule
Negatif & 10,357 & 33.0\% \\
Netral & 17,315 & 55.2\% \\
Positif & 3,688 & 11.8\% \\
\midrule
\textbf{Total} & \textbf{31,360} & \textbf{100\%} \\
\bottomrule
\end{tabular}
\end{table}

The distribution is naturally imbalanced, with Netral being the majority class and Positif the minority. This imbalance reflects the real-world distribution of sentiment in news and social media text, where factual reporting (neutral) is more common than opinionated text, and negative sentiment tends to be more prevalent than positive.

An interesting pattern emerges when examining sentiment by the original relevancy label: texts originally labeled as relevant to their context show a markedly different sentiment distribution (54.0\% negative, 24.2\% neutral, 21.9\% positive) compared to texts labeled as not relevant (22.2\% negative, 71.3\% neutral, 6.5\% positive). Relevant texts are more opinionated, while irrelevant texts are predominantly neutral. This confirms that topical engagement correlates with sentiment expression.

\subsection{Training}

Training follows the same protocol as IndoBERT-Relevancy. We train for 5 epochs with learning rate $2 \times 10^{-5}$, batch size 16, maximum sequence length 256 tokens, and early stopping with patience 2 based on F1 macro on a 15\% stratified validation set. We use inverse-frequency class weighting (Negatif: 1.009, Netral: 0.604, Positif: 2.834) to address the class imbalance. Training was conducted on a single NVIDIA RTX 3090 GPU and completed in approximately 30 minutes.

\section{Baseline Models}

We compare against the three most downloaded general-purpose Indonesian sentiment classifiers available on HuggingFace, all of which are fine-tuned on the SmSA dataset from IndoNLU:

\begin{enumerate}
    \item \textbf{BERT-Indonesian-SmSA} --- BERT Base Indonesian (124M parameters), approximately 108,000 downloads. The most downloaded Indonesian sentiment model.
    \item \textbf{RoBERTa-Indonesian-Sentiment} --- RoBERTa Base Indonesian (124M parameters), approximately 80,000 downloads. Uses a different pre-training approach (RoBERTa) but the same fine-tuning data.
    \item \textbf{IndoBERT-Sentiment-SmSA} --- IndoBERT Base P1 (110M parameters), approximately 15,000 downloads. Uses the same IndoBERT family as our model but the Base variant (110M vs our 335M).
\end{enumerate}

All three models take a single text as input and produce a three-class sentiment prediction without any topical context. They represent the current standard approach to Indonesian sentiment analysis.

\section{Results}

\subsection{Overall Performance}

Table~\ref{tab:results} presents the head-to-head comparison on our held-out test set of 4,704 samples.

\begin{table}[H]
\centering
\caption{Performance comparison on the same test set (4,704 samples). All baseline models are general-purpose (no context). Best results in bold.}
\label{tab:results}
\begin{tabular}{llcccc}
\toprule
\textbf{Model} & \textbf{Type} & \textbf{Accuracy} & \textbf{F1 Macro} & \textbf{F1 Wtd.} & \textbf{Params} \\
\midrule
\textbf{IndoBERT-Sentiment (ours)} & context & \textbf{88.1\%} & \textbf{0.856} & \textbf{0.880} & 335M \\
IndoBERT-Sentiment-SmSA & general & 62.8\% & 0.487 & 0.612 & 110M \\
BERT-Indonesian-SmSA & general & 62.1\% & 0.486 & 0.607 & 124M \\
RoBERTa-Indonesian-Sentiment & general & 59.1\% & 0.501 & 0.593 & 124M \\
\bottomrule
\end{tabular}
\end{table}

Our model outperforms the best baseline by 25.3 percentage points in accuracy and 35.6 points in F1 macro. The improvement is consistent across all metrics.

\begin{figure}[H]
\centering
\begin{tikzpicture}
\begin{axis}[
    ybar,
    ylabel={Score},
    symbolic x coords={Accuracy,F1 Macro,F1 Weighted},
    xtick=data,
    ymin=0.35,
    ymax=1.0,
    bar width=10pt,
    legend style={at={(0.5,-0.2)}, anchor=north, legend columns=2, font=\small},
    width=0.9\columnwidth,
    height=6.5cm,
    enlarge x limits=0.25,
    nodes near coords,
    nodes near coords style={font=\tiny},
    every node near coord/.append style={above},
]
\addplot[fill=blue!60] coordinates {(Accuracy,0.881) (F1 Macro,0.856) (F1 Weighted,0.880)};
\addplot[fill=red!40] coordinates {(Accuracy,0.628) (F1 Macro,0.487) (F1 Weighted,0.612)};
\addplot[fill=green!40] coordinates {(Accuracy,0.621) (F1 Macro,0.486) (F1 Weighted,0.607)};
\addplot[fill=orange!40] coordinates {(Accuracy,0.591) (F1 Macro,0.501) (F1 Weighted,0.593)};
\legend{Ours (context), IndoBERT-SmSA, BERT-Indo-SmSA, RoBERTa-Indo}
\end{axis}
\end{tikzpicture}
\caption{Overall performance comparison. Our context-conditioned model (blue) substantially outperforms all three general-purpose baselines.}
\label{fig:overall}
\end{figure}
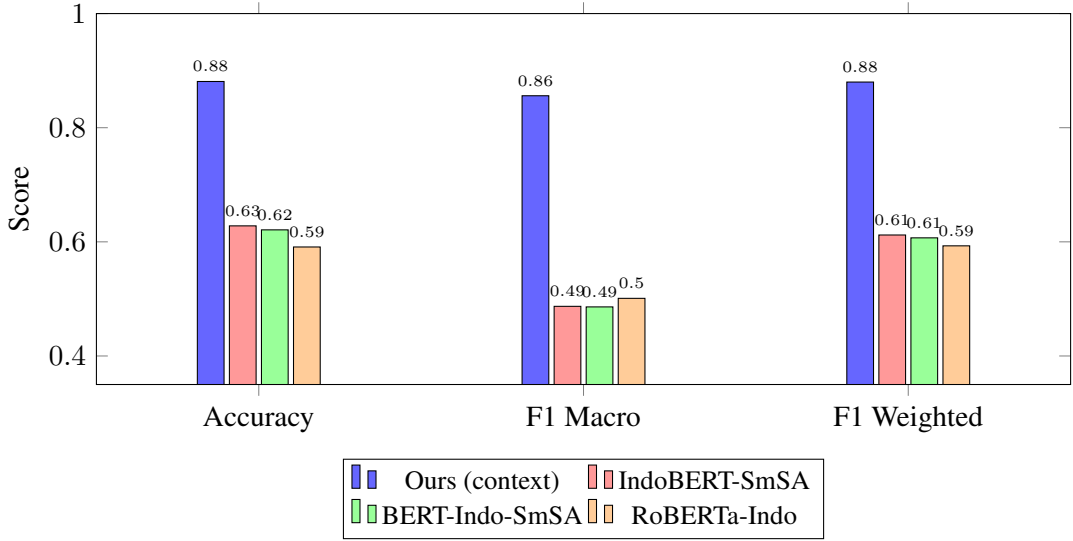

\subsection{Per-Class Performance}

The performance gap is not uniform across classes. Table~\ref{tab:perclass} shows the per-class F1 scores.

\begin{table}[H]
\centering
\caption{Per-class F1 scores. The improvement is most dramatic for the Positif class.}
\label{tab:perclass}
\begin{tabular}{lcccc}
\toprule
\textbf{Class} & \textbf{Ours} & \textbf{IndoBERT-SmSA} & \textbf{BERT-Indo} & \textbf{RoBERTa-Indo} \\
\midrule
Negatif & \textbf{0.876} & 0.608 & 0.606 & 0.654 \\
Netral & \textbf{0.902} & 0.716 & 0.706 & 0.637 \\
Positif & \textbf{0.791} & 0.135 & 0.145 & 0.211 \\
\bottomrule
\end{tabular}
\end{table}

\begin{figure}[H]
\centering
\begin{tikzpicture}
\begin{axis}[
    ybar,
    ylabel={F1 Score},
    symbolic x coords={Negatif,Netral,Positif},
    xtick=data,
    ymin=0,
    ymax=1.0,
    bar width=12pt,
    legend style={at={(0.5,-0.2)}, anchor=north, legend columns=2, font=\small},
    width=0.85\columnwidth,
    height=6cm,
    enlarge x limits=0.3,
    nodes near coords,
    nodes near coords style={font=\tiny},
    every node near coord/.append style={above},
]
\addplot[fill=blue!60] coordinates {(Negatif,0.876) (Netral,0.902) (Positif,0.791)};
\addplot[fill=red!40] coordinates {(Negatif,0.608) (Netral,0.716) (Positif,0.135)};
\addplot[fill=green!40] coordinates {(Negatif,0.606) (Netral,0.706) (Positif,0.145)};
\addplot[fill=orange!40] coordinates {(Negatif,0.654) (Netral,0.637) (Positif,0.211)};
\legend{Ours (context), IndoBERT-SmSA, BERT-Indo-SmSA, RoBERTa-Indo}
\end{axis}
\end{tikzpicture}
\caption{Per-class F1 scores. Baseline models nearly fail on Positif (F1 $\leq$ 0.211), while our model achieves 0.791.}
\label{fig:perclass}
\end{figure}
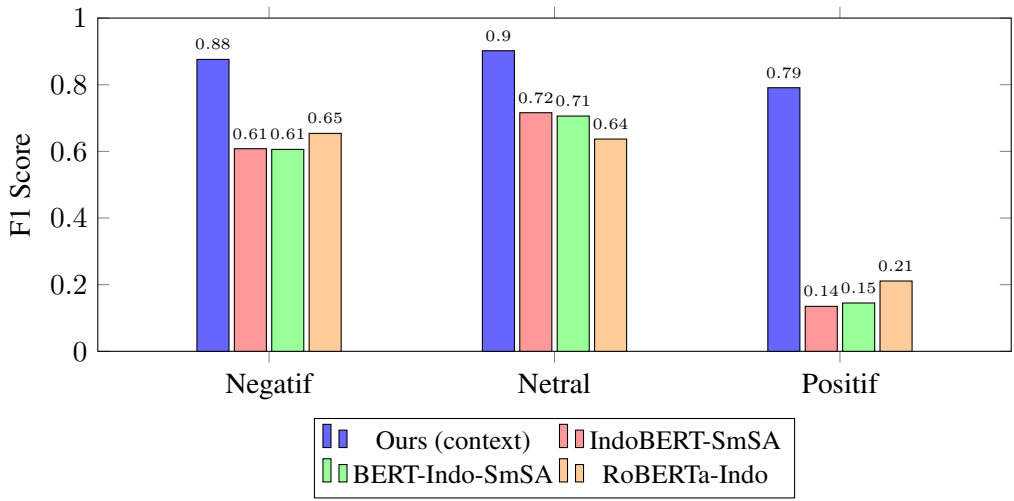

The most striking finding is the collapse of baseline models on the Positif class: all three achieve F1 scores below 0.211, meaning they correctly identify fewer than one in five positive texts. Our model achieves 0.791, a relative improvement of over 275\%.

This failure pattern is revealing. The baseline models were trained on product reviews, where positive sentiment is expressed through words like ``bagus,'' ``suka,'' and ``recommended.'' In news and social media text about public issues, positive sentiment is expressed differently: ``berhasil,'' ``tumbuh,'' ``diresmikan,'' ``ditangkap'' (in the context of law enforcement). Without topical context, these words do not carry obvious positive valence, and the baseline models default to neutral.

\subsection{Qualitative Examples}

Table~\ref{tab:examples} shows representative examples where our model and the baseline models disagree, illustrating the types of texts that benefit from context-conditioning.

\begin{table}[H]
\centering
\caption{Examples where context-conditioning produces correct predictions that all baseline models miss. In each case, the context provides the interpretive frame that determines sentiment.}
\label{tab:examples}
\small
\begin{tabular}{p{3cm}p{5.5cm}ccc}
\toprule
\textbf{Context} & \textbf{Text} & \textbf{Truth} & \textbf{Ours} & \textbf{Baselines} \\
\midrule
Pertumbuhan ekonomi & ekonomi Indonesia tumbuh 5.2\%, tertinggi di ASEAN & Pos & Pos & Net \\
\addlinespace
Inflasi dan daya beli & indomie sekarang 3500, dulu cuma 1500 & Neg & Neg & Net \\
\addlinespace
Korupsi dan penegakan hukum & KPK tangkap bupati yang korupsi dana bansos, akhirnya & Pos & Pos & Net \\
\addlinespace
Kebakaran hutan & Luas Kebakaran Hutan Turun 80\%, Upaya Pencegahan Berhasil & Pos & Pos & Net \\
\addlinespace
Polusi udara & Jakarta peringkat 1 kota paling berpolusi di dunia & Neg & Neg & Pos \\
\addlinespace
Kasus DBD & Kemenkes catat kasus DBD naik 200\% & Neg & Neg & Net \\
\addlinespace
Peredaran narkoba & BNN gagalkan penyelundupan 1 ton sabu dari Malaysia & Pos & Pos & Net \\
\bottomrule
\end{tabular}
\end{table}

These examples share a common pattern: the text contains factual information that is sentiment-bearing \emph{only when interpreted through the lens of the topic}. The arrest of a corrupt official is positive for anti-corruption efforts. A decline in forest fires is positive for environmental protection. A 200\% increase in dengue cases is negative for public health. Without the context, these are simply factual statements, and the baseline models classify them as neutral.

\section{Discussion}

\subsection{Why Context Matters}

The results demonstrate that context-conditioning is not merely a marginal improvement but a qualitative change in capability. The baseline models are not poorly trained; they perform well on the task they were designed for (sentiment classification of product reviews). Their failure on our test set is a domain mismatch problem: news and social media text about public issues expresses sentiment in ways that are fundamentally different from product reviews.

Context-conditioning addresses this mismatch not by training on more diverse sentiment data, but by providing the model with the interpretive frame it needs. The same architecture that learned to determine relevancy can learn to determine sentiment, because both tasks require reasoning about the relationship between a context and a text.

\subsection{Transferability of Context-Conditioned Classification}

This work demonstrates that the context-conditioned classification approach introduced for relevancy \citep{saputra2026indobert} transfers effectively to a different task. The key elements that transferred are:

\begin{enumerate}
    \item \textbf{The architecture}: \texttt{[CLS] context [SEP] text [SEP]} with a classification head.
    \item \textbf{The dataset}: the same 31,360 context--text pairs, re-labeled for the new task.
    \item \textbf{The training protocol}: class weighting, early stopping, stratified validation.
\end{enumerate}

The only change was the labeling: from binary relevancy to three-class sentiment. This suggests that the context-conditioned approach may be applicable to other classification tasks where the meaning of a text depends on an external context, such as stance detection, topic-specific emotion classification, or context-dependent toxicity detection.

\subsection{The Positif Gap}

The most striking result is the failure of baseline models on positive sentiment. All three baselines achieve F1 scores below 0.211 on the Positif class, while achieving reasonable performance on Negatif (0.606--0.654) and Netral (0.637--0.716).

This asymmetry has a simple explanation: negative sentiment in Indonesian public discourse often uses explicitly negative vocabulary (``korupsi,'' ``banjir,'' ``mahal,'' ``macet'') that overlaps with product review vocabulary. Positive sentiment, however, is often expressed through domain-specific achievements (``tumbuh 5.2\%,'' ``berhasil,'' ``diresmikan'') that do not carry positive valence outside their topical context. Without context, these statements are indistinguishable from neutral factual reporting.

\subsection{Binary Variant for Polarity Detection}

While the three-class model (Negatif/Netral/Positif) captures the full spectrum of sentiment, many practical applications require only polarity detection: \emph{is the sentiment positive or negative?} Social media monitoring dashboards, brand reputation tracking, and crisis detection systems often need a clear positive/negative signal without the ambiguity of a neutral class.

To address this, we trained a binary variant of IndoBERT-Sentiment by removing all Netral samples from the training data and re-mapping the labels to Negatif (0) and Positif (1). This yields 14,045 training pairs (10,357 Negatif, 3,688 Positif) across the same 188 topics. The architecture, input format, and training protocol remain identical.

The binary model achieves 96.06\% accuracy and F1 macro of 0.949 on a held-out validation set of 2,107 samples (Table~\ref{tab:binary}). The higher metrics compared to the three-class model reflect the simpler task: without the Netral class, the model only needs to distinguish clear polarity, avoiding the inherently ambiguous boundary between neutral and mildly positive/negative text.

\begin{table}[H]
\centering
\caption{Performance comparison between three-class and binary variants.}
\label{tab:binary}
\begin{tabular}{lccc}
\toprule
\textbf{Variant} & \textbf{Classes} & \textbf{Accuracy} & \textbf{F1 Macro} \\
\midrule
IndoBERT-Sentiment (3-class) & Neg / Net / Pos & 88.1\% & 0.856 \\
IndoBERT-Sentiment (binary) & Neg / Pos & \textbf{96.06\%} & \textbf{0.949} \\
\bottomrule
\end{tabular}
\end{table}

The two variants serve complementary use cases. The three-class model is appropriate when the volume of neutral text is itself informative---for example, when measuring the ratio of opinionated to factual reporting on a topic. The binary model is appropriate when the user has already filtered for opinionated text (e.g., using a relevancy classifier) or simply needs a positive/negative polarity signal. Both models are publicly available on HuggingFace.\footnote{\url{https://huggingface.co/apriandito}}

\subsection{Limitations}

Several limitations should be noted. First, our test set consists of context--text pairs that were labeled using our context-conditioned labeling protocol. This means the ground truth labels inherently reflect a context-dependent notion of sentiment, which may disadvantage context-free models. However, we argue that context-dependent sentiment is the \emph{correct} notion for applications like social media monitoring, where the question is always ``what is the sentiment about topic X?''

Second, our model requires a topical context at inference time, which adds a step compared to context-free models. In practice, this is not burdensome: social media monitoring systems already operate within defined topics.

Third, the class imbalance in our dataset (11.8\% Positif) may affect the absolute performance on the minority class, though class weighting mitigates this substantially.

\section{Conclusion}

We have shown that context-conditioned classification, previously demonstrated for relevancy, transfers effectively to sentiment analysis for Indonesian text. Our model, IndoBERT-Sentiment, achieves F1 macro of 0.856, outperforming the three most widely used Indonesian sentiment models by 35.6 F1 points on the same test set. The improvement is most dramatic for positive sentiment, where context is essential for correct interpretation.

We additionally release a binary variant (Negatif/Positif) that achieves 96.06\% accuracy and F1 macro of 0.949, targeting polarity detection use cases where a neutral class is unnecessary.

The broader implication is that context-conditioning is not task-specific but a general approach to text classification where meaning depends on an external frame of reference. The same architecture and dataset can serve multiple tasks through re-labeling, making it a practical and efficient methodology for building specialized classifiers.

\bibliographystyle{plainnat}

\end{document}